\documentclass[conference]{IEEEtran}
\IEEEoverridecommandlockouts
\usepackage{cite}
\usepackage{amsmath,amssymb,amsfonts}
\usepackage{algorithmic}
\usepackage{graphicx}
\usepackage{textcomp}
\usepackage{xcolor}
\usepackage{enumitem}
\usepackage{xspace}
\usepackage{multirow} 
\usepackage{booktabs} 
\newcommand{\design}{QAgent\xspace}

\usepackage{cite}
\usepackage{url}\usepackage{subcaption}

\def\BibTeX{{\rm B\kern-.05em{\sc i\kern-.025em b}\kern-.08em
    T\kern-.1667em\lower.7ex\hbox{E}\kern-.125emX}}
\begin{document}

\title{\design: An LLM-based Multi-Agent System for Autonomous OpenQASM Programming}

% \author{
% \IEEEauthorblockN{
% Zhenxiao Fu\IEEEauthorrefmark{1},
% Lei Jiang\IEEEauthorrefmark{1},
% Yilun Xu\IEEEauthorrefmark{2},
% Gang Huang\IEEEauthorrefmark{2},
% Fan Chen\IEEEauthorrefmark{1}
% }

% \IEEEauthorblockA{
% \IEEEauthorrefmark{1}Indiana University Bloomington \hspace{3.5em}
% \IEEEauthorrefmark{2}Lawrence Berkeley National Laboratory\\
% \IEEEauthorrefmark{1}\{zhfu, jiang60, fc7\}@iu.edu \hspace{4em}
% \IEEEauthorrefmark{2}\{yilunxu, ghuang\}@lbl.gov}
% }
\author{
\IEEEauthorblockN{
Zhenxiao Fu\IEEEauthorrefmark{1},
Lei Jiang\IEEEauthorrefmark{1},
Yilun Xu\IEEEauthorrefmark{2},
Gang Huang\IEEEauthorrefmark{2},
Fan Chen\IEEEauthorrefmark{1}
}

\IEEEauthorblockA{
\begin{tabular}{@{}l@{\hspace{2.5em}}l@{}}
\IEEEauthorrefmark{1}Indiana University Bloomington
&
\IEEEauthorrefmark{2}Lawrence Berkeley National Laboratory
\\
\IEEEauthorrefmark{1}\{zhfu, jiang60, fc7\}@iu.edu
&
\IEEEauthorrefmark{2}\{yilunxu, ghuang\}@lbl.gov
\end{tabular}
}
}
\maketitle

\begin{abstract}
Programming quantum circuits at the OpenQASM level is essential for achieving hardware-aware optimization and reliable execution on noisy intermediate-scale quantum (NISQ) devices, yet it remains challenging due to the need for domain-specific planning, iterative code synthesis, and low-level calibration. In this paper, we present \design, the first autonomous multi-agent framework for end-to-end OpenQASM code generation. \design\ integrates schema-aware task planning, example- and tool-driven code synthesis, and hardware-aware calibration within a unified planning--synthesis--calibration workflow. The system leverages retrieval-augmented generation (RAG) to access structured kernel knowledge, examples, and backend constraints, and employs coordinated multi-agent reasoning with iterative execution feedback to ensure correctness. We evaluate \design\ on 12 representative quantum kernels and their compositions across five large language models (LLMs). Results show that \design\ improves Pass@1 accuracy by 47--70\% on single-kernel tasks and achieves over 88\% accuracy on multi-kernel workflows for large models, substantially outperforming existing baselines. Furthermore, under realistic hardware frequency drift, \design\ maintains near-unit execution fidelity through automated calibration, whereas SDK-based LLM methods suffer significant degradation. These results demonstrate that integrating planning, synthesis, and calibration is critical for reliable quantum program generation. The implementation of \design\ is open-sourced at \url{https://github.com/fuzhenxiao/QAgent}.
\end{abstract}


\section{Introduction}
\label{s:intro}

Noisy Intermediate-Scale Quantum (NISQ) devices have demonstrated quantum advantage in classically intractable domains, including physical simulation, chemistry, and combinatorial optimization. These capabilities are enabled in large part by the Open Quantum Assembly Language (OpenQASM)~\cite{cross2022openqasm}, which provides a standardized interface between quantum software development kits (SDKs), such as Qiskit~\cite{qiskit2024} and PennyLane~\cite{bergholm2018pennylane}, and heterogeneous quantum hardware platforms~\cite{castelvecchi2023ibm, moses2023race}. While modern SDKs offer high-level abstractions for quantum circuit design and can export subsets of OpenQASM, the generated circuits often omit hardware-specific decompositions, pulse-level optimizations, and calibration data due to limited access to low-level device control. As a result, direct OpenQASM programming remains critical for fine-grained compiler optimization~\cite{litteken2020updated}, experimental calibration and tuning~\cite{mckay2018qiskit}, and hardware-aware pulse-level execution~\cite{khammassi2021openql}.

OpenQASM programming remains challenging, particularly for non-experts, due to its reliance on low-level hardware details such as qubit frequencies, temporal drift, control-port mappings, and pulse scheduling~\cite{amazonbraketexamples,Ramalho:TSEM2025}. To address these complexities, OpenQASM development typically follows a structured, multi-stage workflow~\cite{cross2022openqasm}. In the \textit{planning} stage, high-level algorithmic objectives are decomposed into smaller computational tasks. Suitable circuit kernels~\cite{yang2024qcircuitnet} are then selected and mapped onto available hardware resources, taking into account device constraints and kernel-specific requirements. The subsequent \textit{code synthesis} stage generates the corresponding OpenQASM implementation by defining quantum and classical registers, composing kernel operations, and assembling the full circuit. While canonical algorithms such as Deutsch--Jozsa can be synthesized using well-established templates, more complex or parameterized kernels often require specialized libraries and arithmetic abstractions. This stage typically concludes with systematic verification and debugging. Finally, the \textit{calibration} stage incorporates backend-specific information to mitigate device variability~\cite{mckay2018qiskit,khammassi2021openql}. This involves pulse-level tuning and control adjustments on target qubits to achieve high-fidelity execution. Overall, effective OpenQASM programming demands integrated expertise spanning quantum algorithms, software frameworks, and hardware-level calibration.

Large Language Model (LLM)-based agents~\cite{Tang:NIPS2024,wang2024executable} have demonstrated strong performance in classical code generation across languages such as C, Python, and Verilog~\cite{copilot_agent,nijkamp2023codegen2,zehua2024betterv,ho2025verilogcoder}, frequently leveraging ReAct-style reasoning~\cite{yao2022react} and multi-agent coordination~\cite{zehua2024betterv,ho2025verilogcoder}. In contrast, their application to quantum computing remains comparatively immature. Current efforts primarily rely on static LLMs—pre-trained, fine-tuned, or prompt-engineered—to address narrowly scoped tasks, including quantum physics problem solving~\cite{pan2025quantum,zhou2025application}, circuit construction~\cite{jern2025fine,yang2024qcircuitnet,nakaji2024generative,liang2023unleashing,kashani2024quantumllminstruct,sinha2025circuitpartitioningusinglarge,Aloisio:ESEM2024}, and algorithm explanation~\cite{Aloisio:ESEM2024}. Although such models can generate high-level SDK code (e.g., Qiskit or PennyLane)~\cite{dupuis2024qiskit,basit2025pennylang}, which can be transpiled into OpenQASM, they do not incorporate hardware calibration into the generation process. Moreover, existing LLM-based quantum agents remain confined to domain-specific applications, including quantum chemistry~\cite{zou2025agente}, error correction~\cite{campbell2025enhancing}, and experimental automation~\cite{cao2025agents}. Critically, prior approaches fail to support the integrated, multi-stage workflow—encompassing planning, synthesis, and calibration—required for generating executable, hardware-aware OpenQASM programs.

\begin{figure*}[t]
\centering
\includegraphics[width=0.95\linewidth]{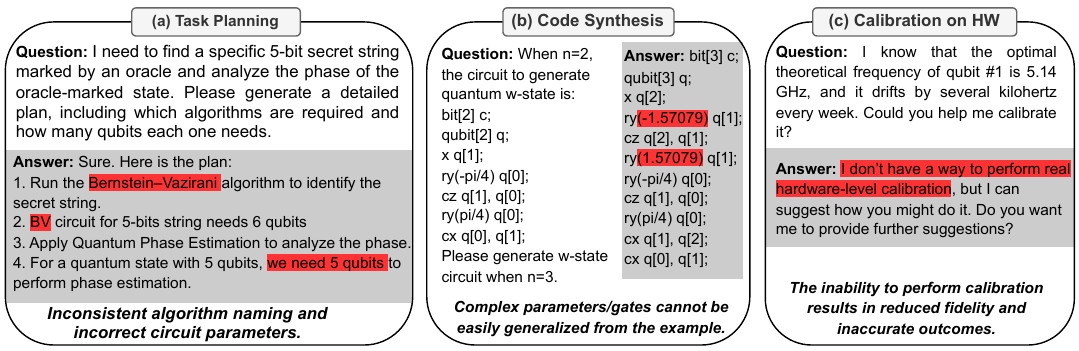}
\vspace{-0.05in}
\caption{Evaluating knowledge-intensive OpenQASM code generation on Qwen235B~\cite{yang2025qwen3technicalreport} reveals substantial limitations across three stages: (a) task planning, (b) code synthesis, and (c) calibration on hardware.}
\label{fig:motivation}
\vspace{-0.1in}
\end{figure*}

Building such an LLM-based agent system poses three key technical challenges that cannot be addressed by standalone LLMs. First, accurate task planning requires domain-specific knowledge—such as kernel input–output specifications and hardware resource constraints—that is typically absent from general-purpose pretraining corpora. Second, synthesizing valid OpenQASM code necessitates iterative generation, execution, reasoning, and debugging, which exceeds the capabilities of single-pass inference. Third, producing hardware-aware programs requires programmatic access to device-specific calibration data (e.g., qubit frequencies, drift characteristics, and pulse schedules), for which LLMs lack native integration mechanisms. Our observations with Qwen3-235B further substantiate these limitations (Figure~\ref{fig:motivation}).

In this paper, we introduce \design, the first autonomous multi-agent framework for low-level OpenQASM code generation. \design\ supports the planning--synthesis--calibration workflow by combining explicit knowledge retrieval via RAG~\cite{lewis2020retrieval} with coordinated multi-agent reasoning, testing, and verification. Our contributions are as follows:
\begin{itemize}[leftmargin=*, topsep=0pt, itemsep=0pt]
    \item \textbf{Schema-aware planning.} We define a structured JSON schema that captures each kernel's functionality, I/O interface, parameters, and resource requirements. A \textit{Planner} decomposes user intents into executable workflows, while a \textit{Criticizer} iteratively checks schema consistency and agent capability constraints.

    \item \textbf{Example- and tool-aware code synthesis.} Based on the resulting plan, coding tasks are assigned to either the \textit{Guided Few-Shot Coding Agent} (GFCA) for structured kernels or the \textit{Tools-Augmented Coding Agent} (TACA) for parameterized or complex kernels. Both agents incorporate an \textit{Advisor} for retrieval, a \textit{Coder} for synthesis, and a \textit{Reviewer} for simulation-based validation.

    \item \textbf{Hardware-aware calibration.} To account for current device conditions, a \textit{Calibration Agent} parses backend constraint files, including qubit parameters, control ports, frequencies, drift, and pulse schedules, or follows expert directives to produce hardware-aligned OpenQASM code.

    \item \textbf{Integrated evaluation.} Evaluated on twelve representative quantum kernels and their compositions across five LLMs, \design\ improves Pass@1 by 47--70\% on single-kernel tasks and achieves over 88\% accuracy for large LLMs on multi-kernel benchmarks, whereas baselines reach only 12--30\% and generally fail on composed tasks. The source code is available at \url{https://github.com/fuzhenxiao/QAgent}.
\end{itemize}

\section{Background and Motivation}

\subsection{OpenQASM Programming}

\textbf{OpenQASM}. Quantum algorithms are ultimately realized as quantum circuits~\cite{nielsen2010quantum}. Although high-level SDKs such as Qiskit~\cite{dupuis2024qiskit} and PennyLane~\cite{basit2025pennylang} can generate OpenQASM as a hardware-level intermediate representation, the resulting circuits often neglect device-specific constraints, e.g., calibration. Consequently, direct OpenQASM programming remains essential for guiding compiler optimizations~\cite{litteken2020updated}, enabling pulse-level refinements~\cite{mckay2018qiskit}, and supporting hardware-aware calibration and control~\cite{khammassi2021openql}, particularly on NISQ devices~\cite{chong2017programming}.

\textbf{Workflow}. OpenQASM programming typically follows a structured workflow~\cite{cross2022openqasm}:
(1) \textit{Task decomposition and kernel selection}. The target algorithm is decomposed into subtasks, and appropriate circuit kernels~\cite{yang2024qcircuitnet} are selected from reference implementations or utility libraries. Standard kernels (e.g., Deutsch--Jozsa) follow canonical templates, whereas complex or parameterized kernels (e.g., quantum adders) often require specialized toolkits, such as register constructors or arithmetic macros.
(2) \textit{Circuit drafting}. Gate sequences are specified, registers are initialized, and quantum and classical resources are allocated.
(3) \textit{Testing and refinement}. The circuit is iteratively validated and debugged to identify and correct logical and syntactic errors.

\textbf{Calibration}. OpenQASM represents quantum programs using abstract operations (e.g., unitary transformations, measurements, and resets), which are ultimately implemented via platform-dependent, time-varying control signals. For example, in superconducting NISQ systems, logical gates are realized via shaped microwave pulses tuned to qubit resonance frequencies. These frequencies can drift by hundreds of kilohertz~\cite{xu2023drift} due to noise and environmental fluctuations~\cite{Clerk:RMP2010}, leading to reduced gate fidelity. To mitigate this, hardware providers (e.g., IBM) perform periodic recalibration and publish updated \textit{constraint files} containing qubit frequencies, coupling maps, pulse parameters, and device limits~\cite{xu2023drift}. OpenQASM~3 incorporates such hardware specificity through calibration constructs, including \texttt{defcal} and \texttt{defcalgrammar}, which bind logical operations to calibrated pulse-level implementations. During compilation, gate invocations are automatically resolved using these updated definitions, ensuring hardware-consistent execution.

\subsection{Current Applications of LLMs in Quantum Tasks}

The success of LLMs in classical domains~\cite{copilot_agent,nijkamp2023codegen2,zehua2024betterv,ho2025verilogcoder} has motivated their application to quantum computing. Existing efforts can be broadly categorized into \textit{static approaches} and \textit{agentic systems}. Static approaches—encompassing pretraining, fine-tuning, and prompt engineering—aim to enhance model capabilities or better exploit latent knowledge. Prior work has improved LLM performance in circuit optimization and architecture design~\cite{jern2025fine,liang2023unleashing}, as well as circuit interpretation~\cite{Aloisio:ESEM2024,kashani2024quantumllminstruct}. Some static LLMs can generate high-level SDK programs using frameworks such as Qiskit and PennyLane~\cite{basit2025pennylang,dupuis2024qiskit}, which can be transpiled into OpenQASM; however, they do not incorporate hardware calibration. QCircuitNet~\cite{yang2024qcircuitnet} further enhances OpenQASM generation by providing large-scale example corpora. In addition, structured prompting techniques enable stepwise reasoning for quantum physics problem solving~\cite{pan2025quantum}. Agentic systems, by contrast, emphasize task decomposition, domain-aware reasoning, and integration with external tools and knowledge sources. For example, EI-Agente~\cite{zou2025agente} analyzes user inputs, selects tools for molecular geometry generation, and returns simulation results, while the framework in~\cite{cao2025agents} employs inspection agents to incorporate external information (e.g., figures) into experimental workflows. These paradigms are complementary rather than mutually exclusive; for instance,~\cite{campbell2025enhancing} combines an agentic framework for Qiskit code generation with fine-tuning on public datasets. Despite this progress, existing approaches are largely confined to domain-specific applications or high-level programming tasks. A fully automated OpenQASM programming pipeline—spanning user intent specification, structured task decomposition, hardware-aware code synthesis, and calibration—remains an open challenge.

\subsection{Design Motivation}

We conduct a preliminary evaluation of Qwen3-235B~\cite{yang2025qwen3technicalreport}, a 235-billion-parameter general-purpose LLM, to assess its capability in OpenQASM code generation. Following recent agentic code-generation paradigms~\cite{zehua2024betterv,ho2025verilogcoder}, we decompose the workflow into three stages—planning, code synthesis, and validation—mirroring the structure of quantum assembly programming. The results, summarized in Figure~\ref{fig:motivation}, indicate that while the model exhibits partial competence, it shows consistent limitations across all stages. In the planning stage, as shown in Figure~\ref{fig:motivation}(a), the model can identify relevant functional kernels (e.g., Bernstein--Vazirani and phase estimation) but fails to allocate resources correctly due to insufficient domain knowledge (e.g., a 5-qubit phase estimation circuit requires an additional ancilla qubit). It also inconsistently refers to the same kernel using different naming variants, leading to ambiguity and potential errors. In the code synthesis stage (Figure~\ref{fig:motivation}(b)), even when provided with reference examples, the model struggles to generalize to larger problem instances (e.g., failing to derive correct parameter values such as \texttt{ry}(-0.955317) and \texttt{ry}(0.955317)). In the validation stage, the model lacks the ability to access or manipulate backend constraint files and therefore cannot perform calibration or adapt programs to hardware-specific conditions. These findings indicate that reliable OpenQASM generation requires an integrated system that combines domain-aware planning with structured kernel representations, supports iterative code synthesis with execution-based feedback, and enables programmatic interaction with hardware calibration data—capabilities not jointly provided by existing LLM-based approaches.

\begin{figure*}[t]
\centering
\includegraphics[width=0.7\linewidth]{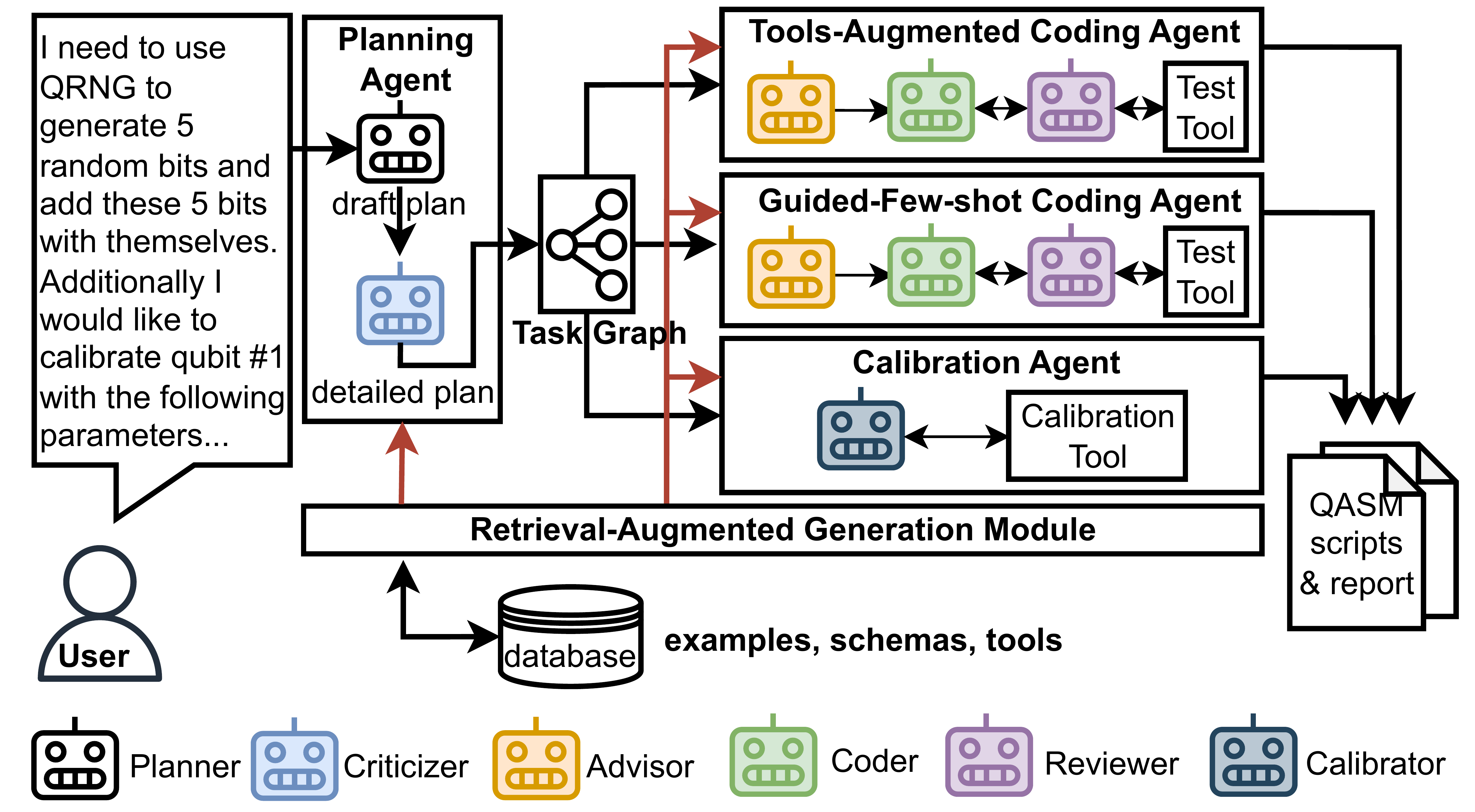}
\vspace{-0.05in}
\caption{The workflow of~\design.}
\label{fig:workflow}
\vspace{-0.1in}
\end{figure*}

\section{\design}

In this section, we present \design, an LLM-based multi-agent framework for OpenQASM code generation that spans multiple coordinated stages, including planning, code synthesis, and calibration. We first provide an overview of the system workflow and then describe each stage in detail. \design\ systematically organizes domain knowledge required for OpenQASM programming, including structured programming schemas, kernel-level code examples, language utilities, tool interfaces, and backend constraint specifications. At each stage, relevant information is dynamically retrieved via Retrieval-Augmented Generation (RAG)~\cite{lewis2020retrieval}. The overall reasoning process is orchestrated using the ReAct paradigm~\cite{yao2022react}, coupled with iterative review and refinement mechanisms. This design enables coordinated multi-agent interaction, improving robustness and reliability in end-to-end code generation.

\subsection{Overall Workflow}

As illustrated in Figure~\ref{fig:workflow}, \design\ processes natural-language OpenQASM requests through three coordinated stages: task planning, code synthesis, and calibration. The \textit{Planning Agent} interprets the user input, retrieves relevant schemas and examples via the RAG module, and decomposes the request into a graph-structured execution plan that specifies individual subtasks and their dependencies. This plan is then passed to the code synthesis stage, where two complementary agents operate: the \textit{Guided Few-Shot Coding Agent} (GFCA), which handles structured kernels using retrieved exemplars, and the \textit{Tools-Augmented Coding Agent} (TACA), which addresses complex or parameterized kernels through specialized utilities. Each agent follows an iterative advisor--coder--reviewer loop to generate, test, and refine subcircuits. The validated components are subsequently integrated into a unified OpenQASM program. Finally, the \textit{Calibration Agent} incorporates backend constraint files or user-provided calibration directives, producing a hardware-aligned OpenQASM program along with an associated calibration report.

\begin{figure*}[t]
\centering
\includegraphics[width=0.8\linewidth]{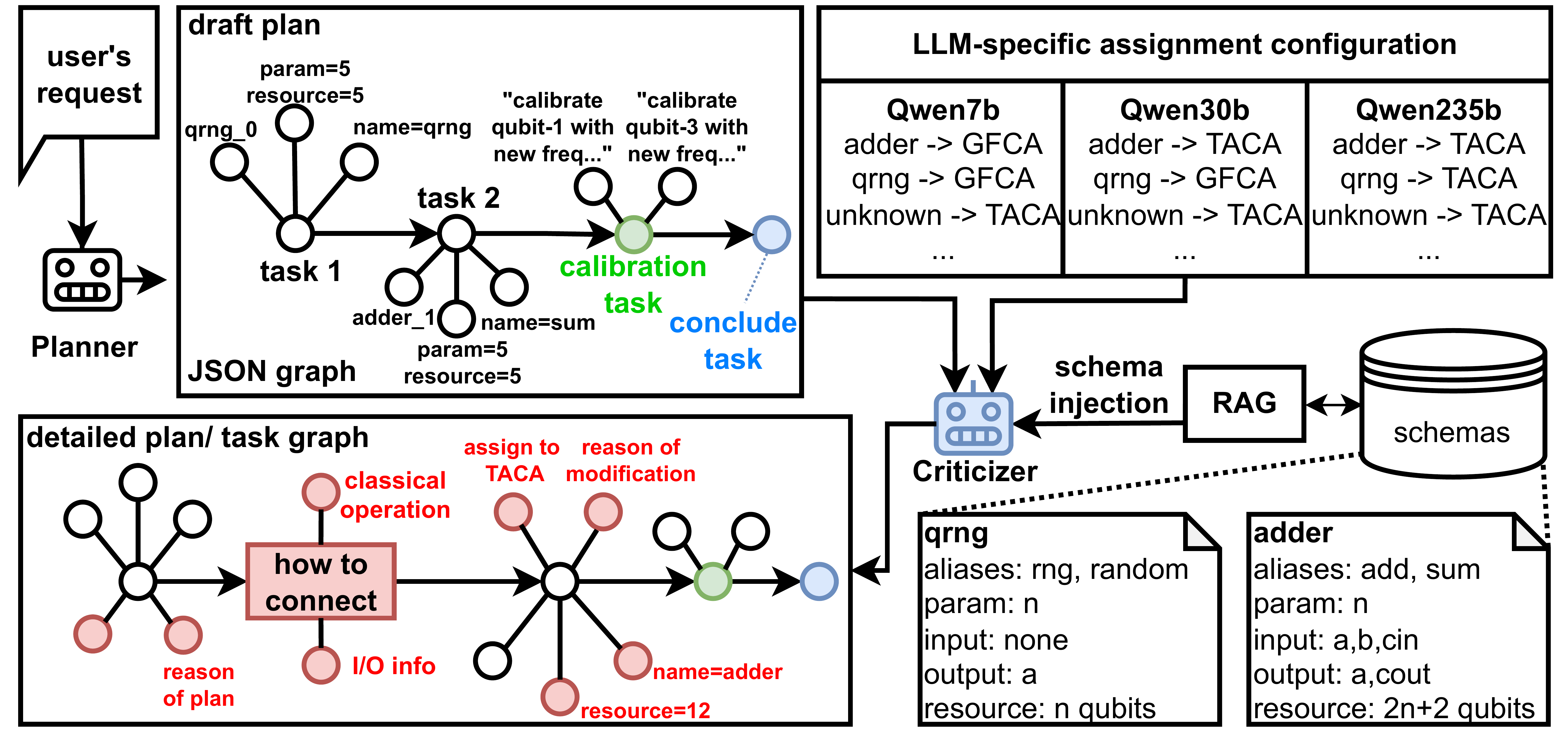}
\vspace{-0.05in}
\caption{An illustration of our Planning Agent.}
\label{fig:plan}
\vspace{-0.2in}
\end{figure*}

\subsection{Task Planning}

The Planning Agent transforms a high-level user instruction into a structured and executable task graph. Within this module, the LLM operates in two complementary roles: a \textit{planner}, which generates an initial draft, and a \textit{criticizer}, which refines and validates the plan into an executable representation (Figure~\ref{fig:plan}). Given a user request, the planner first decomposes it into a set of subtasks, producing a draft workflow that captures the overall structure but may lack accuracy and consistency. As illustrated in Figure~\ref{fig:plan}, a representative request is partitioned into multiple tasks. For example, one task specifies the generation of a quantum random number generator (QRNG) kernel, including its identifier, parameters, and qubit requirements. Another task, corresponding to an adder kernel, highlights typical deficiencies: the planner may assign an incorrect label (e.g., ``sum'') and underestimate required resources (e.g., adding two 5-bit integers requires additional qubits for carry propagation). Additional tasks may include calibration operations, where each node defines the procedure for calibrating individual qubits, and a terminal aggregation step that consolidates outputs. The draft plan is subsequently refined by the criticizer, which integrates two sources of external knowledge via the RAG module: (1) structured schema definitions that specify standardized kernel semantics, including parameters, I/O interfaces, and resource requirements; and (2) predefined routing policies that determine task allocation across agents (e.g., QRNG~$\rightarrow$~GFCA; adder~$\rightarrow$~TACA). By incorporating schema constraints and routing strategies, the criticizer resolves inconsistencies (e.g., correcting kernel names and resource estimates) and augments the plan with additional semantic annotations. These include explicit agent assignments, rationale for task decomposition, and inter-task dependencies expressed through classical control operations (e.g., measurement and state preparation). The resulting task graph is coherent, validated, and semantically complete, serving as the formal specification for downstream execution.

\begin{figure*}[t]
\centering
\includegraphics[width=0.95\linewidth]{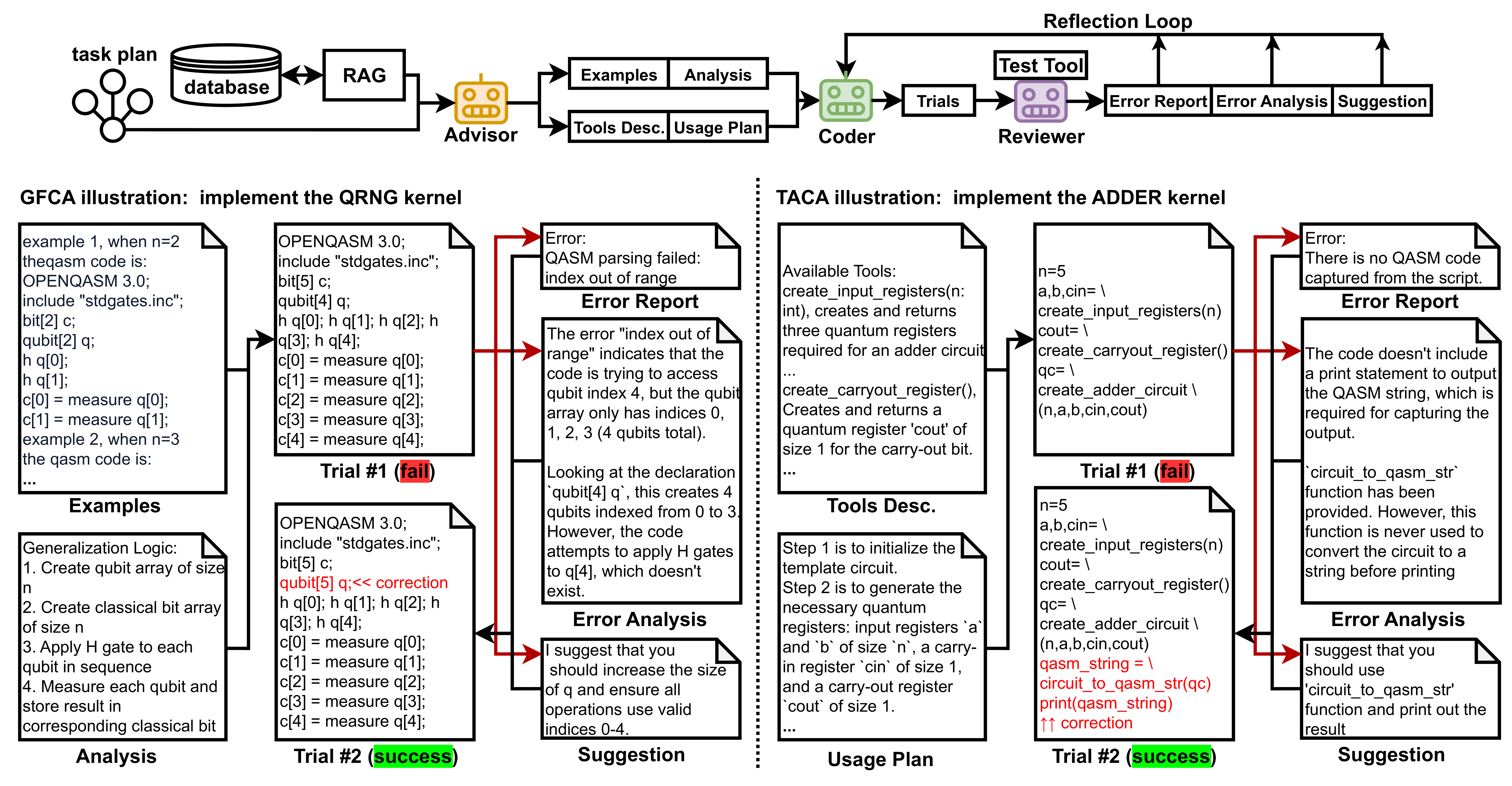}
\vspace{-0.1in}
\caption{An illustration of our Coding Agents.}
\label{fig:gfca}
\vspace{-0.2in}
\end{figure*}

\subsection{OpenQASM Code Synthesis}

Once a task node is dispatched by the Planning Agent, OpenQASM generation proceeds through a unified advisor--coder--reviewer pipeline (Figure~\ref{fig:gfca}). Both the Guided Few-Shot Coding Agent (GFCA) and the Tools-Augmented Coding Agent (TACA) follow this iterative, reflection-driven workflow. The \textit{advisor} retrieves task-relevant knowledge via RAG, the \textit{coder} produces an initial implementation, and the \textit{reviewer} executes the program using testing tools and returns structured feedback, including error messages, diagnostics, and revision suggestions. This closed-loop process iterates until a valid OpenQASM program is obtained. Although GFCA and TACA differ in the type of knowledge they leverage (examples versus tool specifications), the shared reflection mechanism enables robust handling of diverse quantum kernels.

\subsubsection{Guided Few-shot Coding Agent (GFCA)}

GFCA targets regular, pattern-stable kernels using example-driven reasoning. As shown on the left side of Figure~\ref{fig:gfca}, the advisor retrieves multiple annotated examples from the database (e.g., QRNG implementations for various qubit number values of $n$), then analyzes them to infer the general generation logic: create a qubit array, create a classical bit array, apply $H$ gates, measure sequentially, and store results in the corresponding classical bits. Using this distilled structure, the coder drafts an initial OpenQASM program. The reviewer then executes this program and returns a detailed error report. In the illustrated QRNG example, the first trial fails with a QASM parsing error indicating an ``index out of range’’ condition. The error analysis traces the failure to a mismatch between the number of allocated qubits (four) and the indices accessed (0–4), and recommends increasing the register size and ensuring that all measurement operations fall within bounds. The advisor forwards this suggestion back to the coder, who corrects the register declaration and regenerates the QASM code. Trial~2 succeeds, concluding the reflection loop. GFCA thus relies on example-derived structural logic, example-specific errors, and example-aware corrections.

\subsubsection{Tools-Augmented Coding Agent (TACA)}

TACA handles kernels that depend on modular circuit-construction utilities rather than static structural patterns. 
As shown on the right side of Figure~\ref{fig:gfca}, the advisor retrieves database for tool descriptions and function signatures, such as: 
\begin{itemize}
    \item \texttt{create\_input\_registers()}
    \item \texttt{create\_carryout\_register()}
    \item \texttt{create\_adder\_circuit()}
    \item \texttt{circuit\_to\_qasm\_str()}
\end{itemize}
Then, the advisor synthesizes a tools-usage plan outlining the sequence for assembling the adder circuit. The coder follows this plan to generate a Python script that constructs the full quantum adder. The reviewer then executes this script and analyzes both the produced circuit and its QASM output. In the illustrated case, Trial~1 correctly builds the circuit but fails to output any QASM because the coder omitted the call to \texttt{circuit\_to\_qasm\_str()}. The error report identifies that “there is no QASM code captured from the script,” and the error analysis explains that the circuit-to-QASM conversion function must be invoked. The reviewer’s suggestion is passed back to the coder, who adds the missing tool invocation and regenerates the script. Trial~2 succeeds and produces valid QASM. TACA therefore relies on tool catalogs, tool-specific failure modes, and tool-driven corrections.

\subsection{Calibration}

\begin{figure}[t!]
\centering
\includegraphics[width=0.45\textwidth]{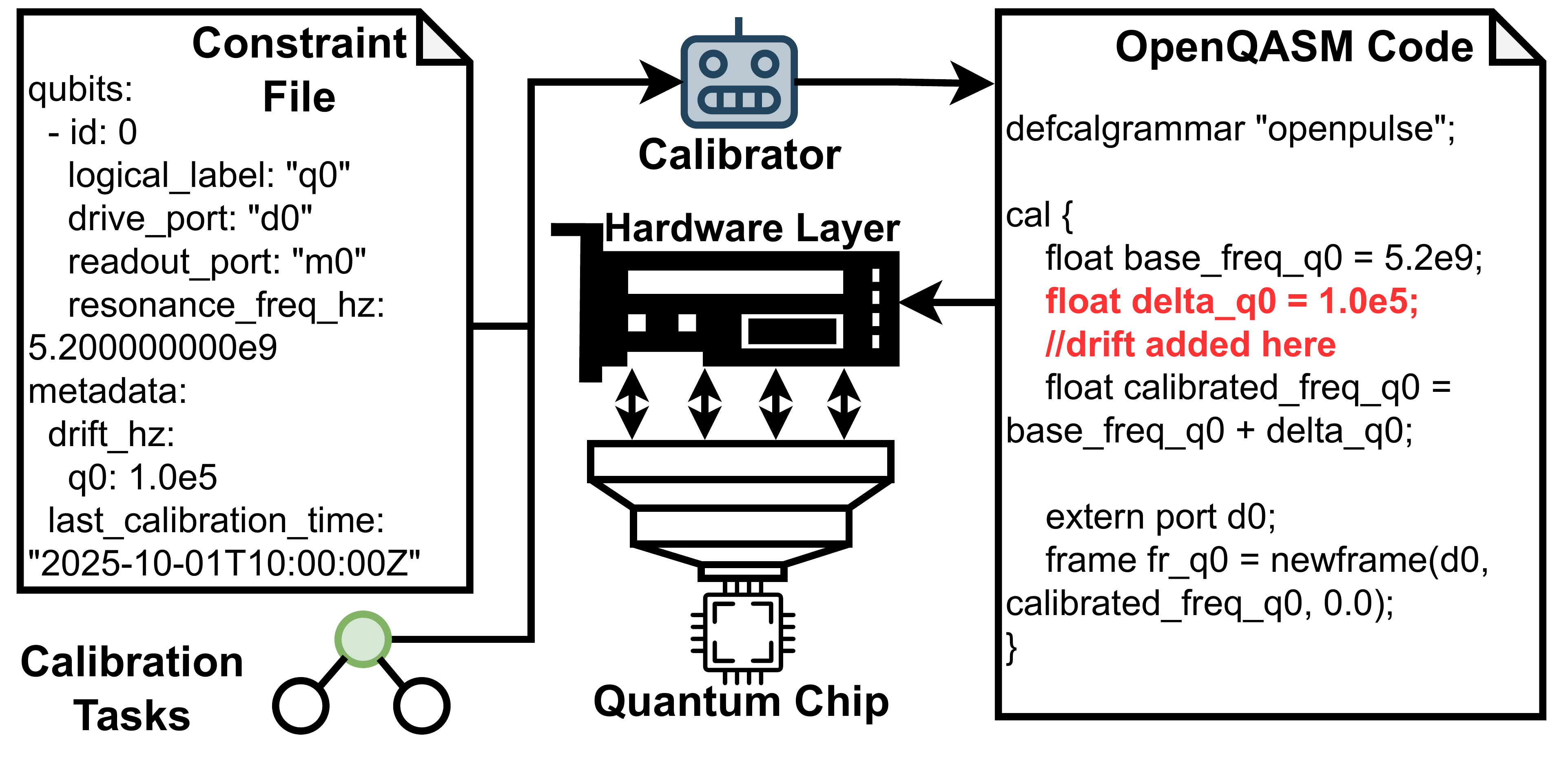}
\caption{An illustration of our Calibration Agent}
\label{fig:cali_agent}
\vspace{-0.1in}
\end{figure}

The Calibration Agent bridges abstract program synthesis and hardware-executable OpenQASM. It operates in two modes depending on available inputs. Expert users may provide explicit calibration directives (e.g., frequency adjustments or control-port mappings), whereas general users rely on backend-supplied constraint files from providers such as IBM. As illustrated in Figure~\ref{fig:cali_agent}, a constraint file specifies per-qubit hardware attributes, including identifiers (e.g., \texttt{id: 0}), logical labels (e.g., \texttt{q0}), control and readout ports (e.g., \texttt{drive\_port: d0} and \texttt{readout\_port: m0}), resonance frequencies, drift parameters (e.g., \texttt{drift\_hz: 0.1e6}), and calibration timestamps. These parameters define the physical operating conditions under which logical OpenQASM instructions are realized at the pulse level. The Calibration Agent parses constraint files to extract parameters relevant to planned calibration tasks. This includes interpreting qubit frequencies, incorporating drift corrections, and establishing mappings between logical qubits and physical control channels. When user-provided calibration directives are available, the agent integrates them by updating or overriding corresponding hardware parameters. Based on this information, the agent generates hardware-aware OpenQASM (or OpenPulse) calibration code, as depicted in Figure~\ref{fig:cali_agent}, where calibrated frequencies, updated drive settings, and device-level specifications are embedded into the program. The resulting output reflects the current hardware state, including drift effects, port mappings, and calibration metadata, ensuring that the synthesized circuit is aligned with device constraints and suitable for reliable execution on the target quantum processor.

% 这个study中用到的12种kernels列出在Table~\ref{t:benchmarks}的最左侧。表中间的数据，作为部分subset的examples，列出了这些kernel在$n = 5$.的时候各自的：简单quantum gate（比如h，x），复杂的custom quantum gate（比如inverse-qft）,还有带有参数的门（比如rx）的数量。

\section{Experimental Methodology}

%%%%%%%%%%%%%%%%%%%%%%%%%%%%%%%%%%%%%%
\begin{table}[t!]
\centering
\caption{Simulated quantum benchmark kernels when $n=5$($n$: qubit count; Q: quantum; Param.: parameterized).}
\label{t:benchmarks}
\setlength{\tabcolsep}{3pt}
\renewcommand{\arraystretch}{1.25}
\footnotesize
\begin{tabular}{|l|l|c|c|c|c|}
\hline
\multicolumn{2}{|c|}{\textbf{Benchmarks}}
& \multicolumn{3}{c|}{\textbf{Number of Gates ($n=5$)}} 
& \multirow{2}{*}{\textbf{Feature}}
\\\cline{1-5}
\textbf{ID} &\multicolumn{1}{|l|}{\textbf{Kernel}}
& \textbf{Simple}& \textbf{Custom}& \textbf{Param.} &\\\hline
1& Bernstein–Vazirani (bv)      & 12& 0& 0 
& Structured /       \\\cline{1-5}

2& Deutsch-Jozsa (dj)           & 12& 0& 0     
& Oracle \\\cline{1-5}
3& Grover's Search (gr)         & 27& 4& 0     
&        \\\hline

4& Adder (ad)                   & 7& 10& 0 
& Composite/ \\\cline{1-5}
5& Or Gate (or)                & 58& 14& 48 
& Parameterized/     \\\cline{1-5}

6& Permutation (pm)             & 4& 1& 0           
& Algorithmic         \\\hline
7& Phase Estimation (pe)        & 22& 1& 0 
&Composite       \\\hline

8& Q RNG (qr)             & 5& 0& 0            
& Measurement\\\hline
9& Q Fourier Transform (qf)  & 20& 0& 36            
& Parameterized \\\hline

10& W-State (ws)                 & 8& 0& 9            
& Multipartite           \\\cline{1-5}
11& GHZ-State (gh)               & 5& 0& 0 
& Entanglement    \\\cline{1-5}
12& Cluster-State (cl)           & 9& 0& 0 
& Preparation    \\ \hline
\end{tabular}
\vspace{-0.1in}
\end{table}
%%%%%%%%%%%%%%%%%%%%%%%%%%%%%%%%%%%%%%

\textbf{Benchmarks}. We evaluate \design\ on a representative set of quantum kernels derived from the QCircuitNet dataset~\cite{yang2024qcircuitnet}, with qubit sizes ranging from 3 to 12. The twelve kernels used in this study are listed in the leftmost column of Table~\ref{t:benchmarks}. To illustrate their characteristics, the central columns report detailed gate counts for a subset of kernels instantiated at a circuit scale of 5 qubits. Specifically, we quantify three categories of operations: basic quantum gates (e.g., H, X), composite or custom gates (e.g., inverse QFT), and parameterized gates (e.g., RX). These kernels span a wide spectrum of circuit complexities, ranging from elementary one- and two-qubit operations to advanced constructions involving custom and parameterized gates that require fine-grained control. This diversity facilitates a comprehensive evaluation of both performance and generalization across heterogeneous circuit structures and workloads. To further assess compositional capability, we construct 15 multi-kernel tasks by combining primitive kernels into more complex quantum workflows, with 5 tasks each for 2-, 3-, and 4-kernel compositions. \textit{Notably, for two kernels—GHZ-State and Cluster-State—the corresponding schemas, examples, and tool definitions are automatically generated by Qwen3-235B. This demonstrates the scalability of \design\ in incorporating new kernels without manual intervention.}

\textbf{LLMs}. We evaluate~\design~using the following LLMs, and use members of the Qwen family with different parameter scales for the ablation study:
\begin{itemize}[leftmargin=*, topsep=0pt, itemsep=0pt]
    \item \textit{Qwen7b}: Qwen2.5-Coder-7B-Instruct~\cite{hui2024qwen25codertechnicalreport},
    \item \textit{Qwen30b}: Qwen3-Coder-30B-A3B-Instruct~\cite{yang2025qwen3technicalreport},
    \item \textit{Qwen235b}: Qwen3-235B-A22B-Instruct-2507~\cite{yang2025qwen3technicalreport},
    \item \textit{Devstral24b}: Devstral-Small-2505~\cite{rastogi2025devstralfinetuninglanguagemodels},
    \item \textit{gpt120b}: gpt-oss-120b~\cite{openai2025gptoss120bgptoss20bmodel}.
\end{itemize}

\textbf{Schemes}. To evaluate \design, we implement and compare the following schemes:
\begin{itemize}[leftmargin=*, topsep=0pt, itemsep=0pt]
\item \textit{Overall performance}. We first evaluate the end-to-end performance of \design\ on both single-kernel and multi-kernel tasks across five LLMs, assessing its ability to complete the full planning--synthesis--calibration workflow as task complexity increases. 

\item \textit{Component-level ablation}. We then analyze the contribution of individual agents within \design. For the Planning Agent, we examine the effect of the proposed schema-aware mechanism on planning accuracy. For the Coding Agents, we compare GFCA and TACA across all twelve kernels against the static baseline~\cite{yang2024qcircuitnet}, which is the closest prior work targeting OpenQASM code generation. For the Calibration Agent, we evaluate circuit fidelity with and without calibration to quantify its impact.

\item \textit{Comparison with SDK-oriented LLM methods}. Finally, we compare \design\ with three representative LLM-based approaches for Qiskit and PennyLane code generation: Qiskit-Code-Assistant (QCA)~\cite{dupuis2024qiskit}, a trained Qiskit-oriented LLM; Pennylang~\cite{basit2025pennylang}, a RAG-based PennyLane generator; and Qiskit-Multi-Agent-System (QMAS)~\cite{campbell2025enhancing}, which integrates fine-tuning with a multi-agent framework for Qiskit programming. As these methods target high-level SDKs rather than OpenQASM directly, their outputs can only be transpiled into limited OpenQASM representations for simple circuits. Accordingly, we restrict this comparison to relatively simple and structured kernels (bv, dj, and qr) to ensure a fair evaluation based on circuit fidelity.
\end{itemize}

% Qwen3-235B-A22B-Instruct-2507 (Qwen235b))\cite{yang2025qwen3technicalreport},
% Qwen3-Coder-30B-A3B-Instruct (Qwen30b)\cite{yang2025qwen3technicalreport},
% Qwen2.5-Coder-7B-Instruct (Qwen7b)\cite{hui2024qwen25codertechnicalreport},
% Devstral-Small-2505 (Devstral24b)\cite{rastogi2025devstralfinetuninglanguagemodels}, 
% gpt-oss-120b (gpt120b)\cite{openai2025gptoss120bgptoss20bmodel}.

\textbf{Evaluation Metric}. We evaluate model performance using the Pass@k metric~\cite{chen2021evaluating}, where Pass@k denotes the probability that at least one of the $k$ independently generated candidate solutions is correct:
\[
\mathrm{pass@}k=
1-\dfrac{\binom{n-c}{k}}{\binom{n}{k}}
\]
Where $n$ is the total number of generated samples for a task and $c$ is the number of correct ones among them. Smaller values of $k$ reflect the model's ability to produce a correct solution within the first few attempts, indicating its per-sample accuracy. We count a generated case as correct only when it is syntactically valid OpenQASM code and its simulated output distribution (1,000 shots) matches the distribution of the golden circuit. Larger values of $k$ instead capture the model's upper-bound performance when more sampling is allowed. In our experiments, 
10 samples are collected from each LLM for each task, and we report results for Pass@1, Pass@3, and Pass@5. In addition, the execution quality of the final generated code is measured by fidelity, a standard metric for assessing quantum circuit performance.

\begin{table}[t!]
\centering
\caption{Overall performance of QAgent with different base LLMs across multi-kernel settings.}
\label{tab:overall-performance}
\setlength{\tabcolsep}{5pt}
\renewcommand{\arraystretch}{1.25}
\footnotesize
\begin{tabular}{|l|l| c| c| c| c|}
\hline
\textbf{Pass@k} & \textbf{LLM} & \textbf{1-kernel} & \textbf{2-kernels} & \textbf{3-kernels} & \textbf{4-kernels} \\\hline
\multirow{5}{*}{Pass@1} 
 & Qwen7b       & 0.59 & 0.21 & 0.03 & 0.02 \\
 & Qwen30b      & 0.97 & 0.89 & 0.77 & 0.73 \\
 & Qwen235b     & 0.98 & 0.94 & 0.94 & 0.88 \\
 & Devstral24b  & 0.87 & 0.65 & 0.34 & 0.26 \\
 & gpt120b      & 0.99 & 1.00 & 0.98 & 1.00 \\
\hline
\multirow{5}{*}{Pass@3} 
 & Qwen7b       & 0.78 & 0.39 & 0.06 & 0.04 \\
 & Qwen30b      & 0.99 & 0.93 & 0.76 & 0.75 \\
 & Qwen235b     & 0.99 & 0.96 & 0.94 & 0.90 \\
 & Devstral24b  & 0.97 & 0.81 & 0.50 & 0.44 \\
 & gpt120b      & 1.00 & 1.00 & 0.98 & 1.00 \\
\hline
\multirow{5}{*}{Pass@5} 
 & Qwen7b       & 0.83 & 0.46 & 0.12 & 0.07 \\
 & Qwen30b      & 1.00 & 0.96 & 0.80 & 0.80 \\
 & Qwen235b     & 1.00 & 0.98 & 0.94 & 0.92 \\
 & Devstral24b  & 0.99 & 0.82 & 0.52 & 0.46 \\
 & gpt120b      & 1.00 & 1.00 & 0.98 & 1.00 \\
\hline
\end{tabular}
\end{table}

\begin{table}[t!]
\centering

\caption{Pass@1 results for the Planning Agents.}
\label{tab:schema_injection}

\setlength{\tabcolsep}{8pt}
\renewcommand{\arraystretch}{1.25}
\footnotesize
\begin{tabular}{|l|c|c|c|}
\hline
\multicolumn{1}{|c|}{\textbf{Scheme}}
& \textbf{Qwen7b} & \textbf{Qwen30b} & \textbf{Qwen235b} \\
\hline
w/o Schema  &0.34  & 0.69 &0.75\\
w/ Schema   
&0.35 (\color{red}{2.9\%$\uparrow$})  
& 0.85 (\color{red}{23.2\%$\uparrow$})
&0.94 (\color{red}{25.3\%$\uparrow$}) \\
\hline
\end{tabular}
\end{table}

\section{Experimental Results}

\subsection{Overall Performance}

Table~\ref{tab:overall-performance} reports the performance of \design\ across different base LLM scales and task complexities, measured by the number of composed quantum kernels. 

We first observe that \design\ with \textbf{gpt120b} achieves near-perfect accuracy across all kernel counts and Pass@$k$ settings, with minimal variance, effectively establishing an empirical upper bound. \textbf{Devstral24b} also performs strongly on single-kernel tasks, reaching 87\% Pass@1 and 99\% by Pass@5. These results indicate that large-scale models can fully leverage the multi-agent workflow—including planning, decomposition, retrieval, iterative refinement, and calibration—to maintain high correctness even for complex compositions.

Within the \textbf{Qwen} family, three key trends emerge.  
(1) \textit{Model scale correlates with multi-kernel robustness.} Qwen7B achieves 59\% Pass@1 on single-kernel tasks but degrades sharply with increasing complexity, dropping to 2\% for four-kernel compositions. Qwen30B shows substantial improvement, achieving 97--100\% on single kernels and 73--96\% on multi-kernel tasks. Qwen235B is the most stable, maintaining 98--100\% on single kernels and strong multi-kernel performance (above 94\% for three kernels and above 88\% for four kernels).  
(2) \textit{Larger models exhibit higher sample efficiency.} Increasing $k$ (Pass@3, Pass@5) yields the largest relative gains for Qwen7B, indicating reliance on sampling to compensate for weaker reasoning. In contrast, Qwen30B and Qwen235B achieve high accuracy with limited sampling.  
(3) \textit{Planning robustness is the primary limitation for smaller models.} The sharp performance decline of Qwen7B in multi-kernel settings reflects difficulty in decomposing and coordinating multi-stage workflows, whereas larger models demonstrate more reliable multi-step reasoning and stronger alignment between planning and synthesis.

Overall, \design\ consistently improves OpenQASM generation across LLM scales. However, sustained high accuracy on multi-kernel tasks depends on the reasoning capacity of larger models. While smaller models benefit from increased sampling, reliable multi-kernel synthesis requires higher-capability backbones.

\subsection{Component-Level Ablation Study}

\textit{Planning Agent.} Table~\ref{tab:schema_injection} reports Pass@1 planning accuracy with and without the proposed schema-awareness mechanism across model scales. Two observations emerge.  
(1) Larger models exhibit stronger baseline planning capability: Qwen7B achieves 34\%, compared to 69\% for Qwen30B and 75\% for Qwen235B, reflecting improved intrinsic ability for task decomposition and workflow reasoning.  
(2) Schema awareness consistently improves accuracy across all models. The gains are most significant for Qwen30B and Qwen235B, with relative improvements of 23.2\% and 25.3\%, respectively, indicating that structured kernel metadata enhances plan consistency and correctness. In contrast, Qwen7B shows only marginal improvement (2.9\%), suggesting limited capacity to utilize structured inputs effectively.

%%%%%%%%%%%%%%%%%%%%%%%%%%%%%%%%%%%%%%%%%
\begin{table*}[t]
  \centering
  \small
  \setlength{\tabcolsep}{1.6pt}
  \renewcommand{\arraystretch}{1.15}
  \caption{Pass@1 results comparison for the Coding Agents.}
  \label{tab:qwen_pass1_schemes}
  \vspace{-0.07in}
  \begin{tabular*}{\textwidth}{@{\extracolsep{\fill}}|l|l|c|c|c|c|c|c|c|c|c|c|c|c|c|}
    \hline
    \textbf{LLM} & \textbf{Scheme} & \textbf{bv} & \textbf{dj} & \textbf{gr} & \textbf{ad} & \textbf{or} & \textbf{pm} & \textbf{pe} & \textbf{qr} & \textbf{qf} & \textbf{ws} & \textbf{gh} & \textbf{cl} & \textbf{avg} \\\hline

    \multirow{3}{*}{Qwen7b}
      & Static & 0.56 & 0.24 & 0.32 & 0.00 & 0.00 & 0.00 & 0.32 & 0.00 & 0.00 & 0.00 & 0.00 & 0.00 & 0.12 \\
      & GFCA   & \textcolor{red}{0.76} & \textcolor{red}{0.60} & \textcolor{red}{0.36} & 0.12 & 0.00 & \textcolor{red}{0.28} & \textcolor{red}{0.32} & \textcolor{red}{0.96} & \textcolor{red}{0.96} & 0.00 & \textcolor{red}{1.00} & \textcolor{red}{0.92} & \textcolor{red}{0.52} \\
      & TACA   & 0.04 & 0.16 & 0.00 & \textcolor{red}{0.36} & \textcolor{red}{0.16} & 0.20 & 0.16 & 0.32 & 0.60 & \textcolor{red}{0.44} & 0.20 & 0.04 & 0.22 \\
    \hline

    \multirow{3}{*}{Qwen30b}
      & Static & 0.80 & 0.88 & 0.40 & 0.00 & 0.00 & 0.00 & 0.64 & 0.20 & 0.00 & 0.00 & 0.00 & 0.36 & 0.27 \\
      & GFCA   & \textcolor{red}{1.00} & \textcolor{red}{1.00} & \textcolor{red}{1.00} & 0.08 & 0.64 & \textcolor{red}{1.00} & 0.56 & \textcolor{red}{1.00} & \textcolor{red}{1.00} & 0.00 & \textcolor{red}{1.00} & \textcolor{red}{1.00} & 0.77 \\
      & TACA   & 0.40 & 0.72 & \textcolor{red}{1.00} & \textcolor{red}{0.92} & \textcolor{red}{1.00} & \textcolor{red}{1.00} & \textcolor{red}{0.92} & 0.96 & \textcolor{red}{1.00} & \textcolor{red}{0.80} & \textcolor{red}{1.00} & 0.72 & \textcolor{red}{0.87} \\
    \hline

    \multirow{3}{*}{Qwen235b}
      & Static & 0.88 & 0.92 & 0.44 & 0.04 & 0.08 & 0.04 & 0.72 & 0.00 & 0.04 & 0.00 & 0.00 & 0.40 & 0.30 \\
      & GFCA   & \textcolor{red}{1.00} & \textcolor{red}{1.00} & 0.52 & 0.72 & 0.52 & \textcolor{red}{1.00} & 0.88 & 0.96 & \textcolor{red}{1.00} & 0.12 & \textcolor{red}{1.00} & \textcolor{red}{1.00} & 0.81 \\
      & TACA   & \textcolor{red}{1.00} & 0.92 & \textcolor{red}{1.00} & \textcolor{red}{1.00} & \textcolor{red}{1.00} & \textcolor{red}{1.00} & \textcolor{red}{1.00} & \textcolor{red}{1.00} & 0.96 & \textcolor{red}{0.80} & 0.96 & 0.92 & \textcolor{red}{0.96} \\
    \hline
  \end{tabular*}
\vspace{-0.1in}
\end{table*}
%%%%%%%%%%%%%%%%%%%%%%%%%%%%%%%%%%%%%%%%%

\begin{figure}
% \vspace{-0.2in}
\begin{subfigure}{.5\textwidth}
  \centering
  \includegraphics[width=1\linewidth]{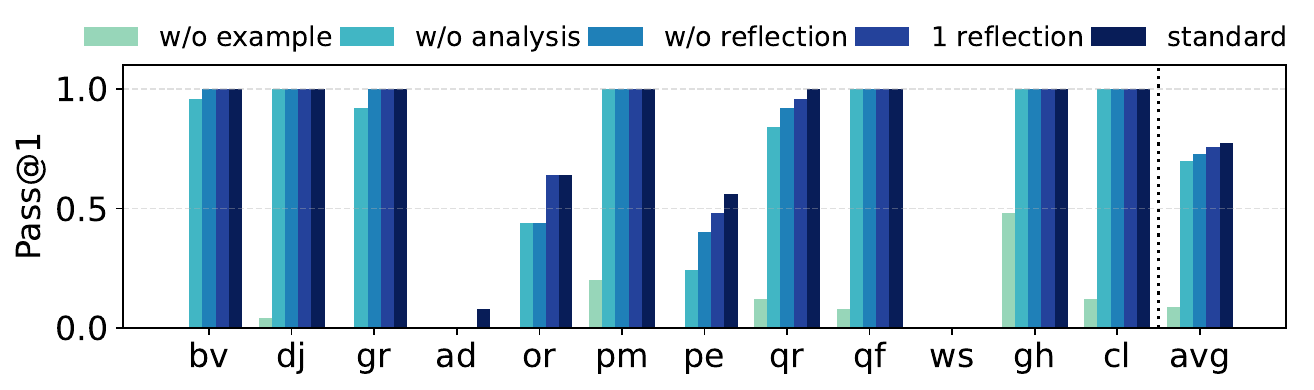} 

  \caption{Pass@1 results across different GFCA configurations.}
  \label{fig:GFCA-ablation}
\end{subfigure}
\begin{subfigure}{.5\textwidth}
  \centering
  \includegraphics[width=1\linewidth]{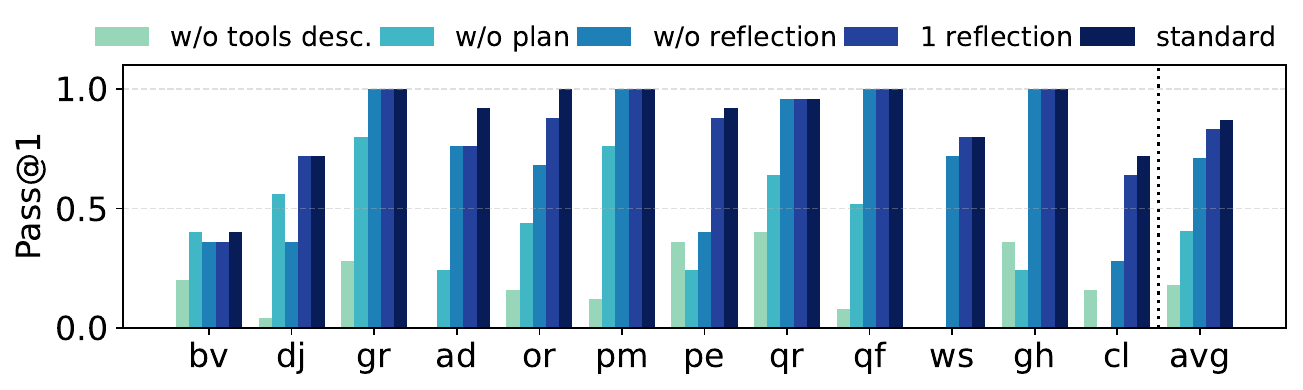}  
  \caption{Pass@1 results across different TACA configurations.}
  \label{fig:TACA-ablation}
\end{subfigure}

\caption{Ablation study of the two Coding Agents.}
\label{fig:coders-ablation}

\end{figure}
%%%%%%%%%%%%%%%%%%%%%%%%%%%%%%%%%%%%%%%%%

\textit{Coding Agents.} Table~\ref{tab:qwen_pass1_schemes} summarizes Pass@1 performance for three coding strategies: Static prompting~\cite{yang2024qcircuitnet}, GFCA-only, and TACA-only. Two conclusions follow. First, static prompting is insufficient for reliable OpenQASM generation, particularly for multi-step or tool-dependent kernels. Second, both GFCA and TACA substantially improve correctness: GFCA performs best on structured kernels, while TACA achieves superior results on complex or tool-intensive kernels, especially for larger models. Strong performance on \texttt{gh} and \texttt{cl} further indicates that \design\ generalizes to kernels whose schemas and tools are LLM-generated.

\textit{GFCA Ablation.} Figure~\ref{fig:GFCA-ablation} evaluates key components of GFCA.  
(1) Removing example retrieval reduces average accuracy from 77.3\% to 8.7\%, demonstrating that representative examples are essential for capturing recurring structural patterns (e.g., \texttt{bv}, \texttt{dj}, \texttt{qr}).  
(2) Removing example analysis decreases accuracy to 70.0\%, indicating that abstracting shared structural logic is necessary for generalization.  
(3) Removing reflection lowers performance to 73.0\%, while limiting to a single iteration yields only 75.7\%, confirming that iterative refinement is important for correcting subtle logical and syntactic errors.  
These trends are consistent across kernels and align with overall performance results.

\textit{TACA Ablation.} Figure~\ref{fig:TACA-ablation} highlights the importance of TACA components.  
(1) Removing tool descriptions reduces accuracy to 18.0\%, showing that explicit API-level knowledge is critical for tool-dependent kernels (e.g., \texttt{ad}, \texttt{or}, \texttt{pm}, \texttt{pe}).  
(2) Removing the usage plan results in 40.3\% accuracy, underscoring the need for structured orchestration of tool calls.  
(3) Removing reflection decreases accuracy from 87.0\% to 71.0\%, and limiting to a single iteration (83.3\%) still underperforms the full configuration, indicating that iterative validation is essential for resolving execution and I/O issues.  
These findings are consistent with Table~\ref{tab:qwen_pass1_schemes}, where TACA outperforms GFCA on tool-intensive kernels.

\textit{Summary on Coding Agents.} Overall, the combined evidence from the performance table and the ablation studies validates the complementary design of the two agents. GFCA is driven by example abstraction and structural pattern inference, enabling strong performance on regular kernels. TACA depends on tool-aware retrieval, procedural planning, and iterative correction, making it well-suited for complex or parameterized kernels. Reflection consistently enhances robustness in both agents, confirming its role as a key mechanism for reliable OpenQASM generation within \design.

\begin{figure}
    \centering
    \includegraphics[width=1\linewidth]{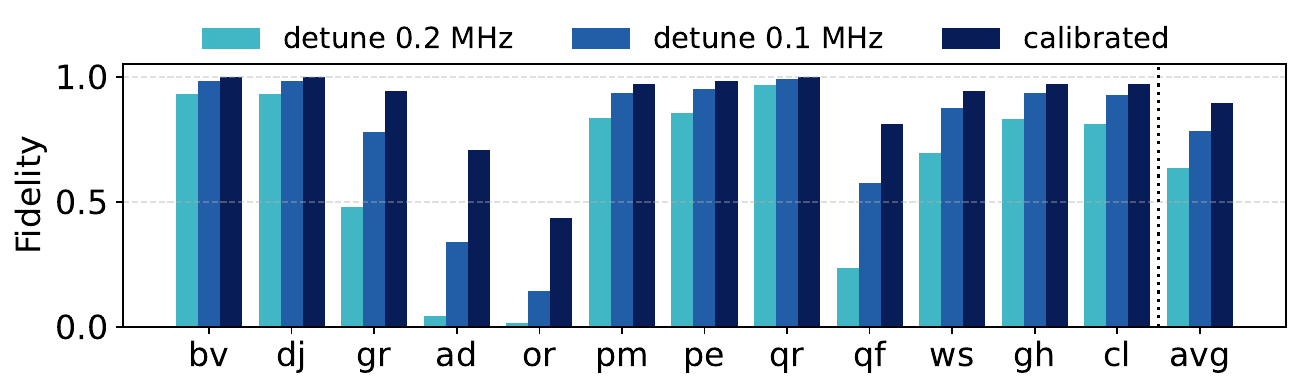}
    \caption{Fidelity performance of the Calibration Agent.}
    \label{fig:cali}
\end{figure}

\textit{Calibration Agent}. We simulate realistic hardware detuning scenarios on superconducting backends in which each qubit’s resonance frequency is shifted by 0.1–0.2~MHz from its nominal value. Such frequency drift is common in NISQ-era devices and, if uncorrected, can significantly reduce gate fidelity and overall circuit correctness~\cite{xu2023drift}. Figure~\ref{fig:cali} reports the resulting circuit fidelities across the 12 benchmark kernels. Under a 0.2~MHz detuning, fidelity degrades substantially, dropping to an average of 63\%. Kernels with greater two-qubit gate depth (such as Grover and Adder) are particularly affected, falling to roughly 48\% and 4\%, respectively. Even with a milder 0.1~MHz drift, the average fidelity remains only 78\%, showing that small but persistent hardware deviations can meaningfully impair execution reliability. Once the Calibration Agent parses the hardware constraint file and inserts calibrated frequency parameters into the generated OpenQASM/OpenPulse program, execution fidelity rebounds across all kernels. The calibrated configuration achieves an average fidelity of 89\%, with high-depth and parameter-sensitive circuits exhibiting improvements exceeding 30 percentage points. These results demonstrate that automated, hardware-aware calibration is essential for mitigating frequency drift and preserving reliable quantum program execution.

\begin{figure}
    \centering
    \includegraphics[width=1\linewidth]{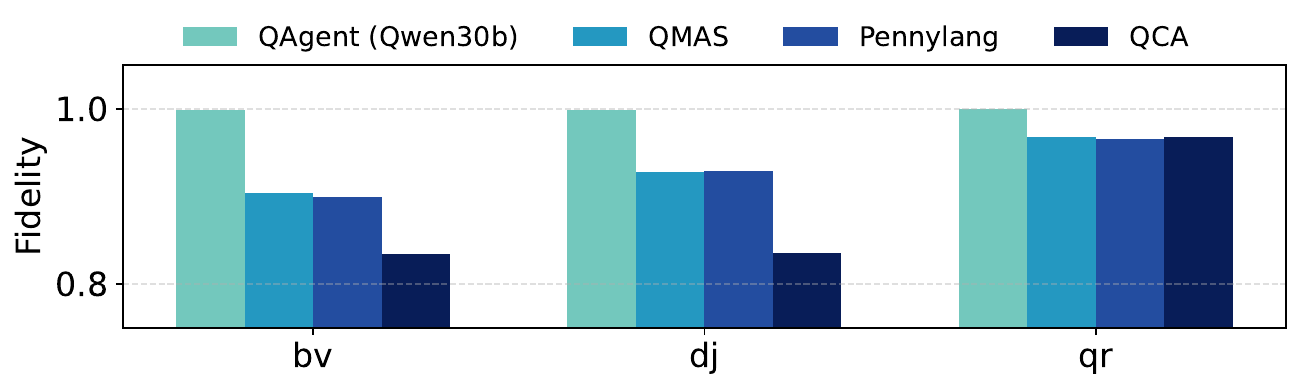}
    \caption{Average fidelity comparison with SDK-oriented LLM methods under 0.2\,MHz frequency detuning (10 attempts, 1000 shots each).}
    \label{fig:work_compare}
\end{figure}

\subsection{Comparison against SDK-oriented LLMs}

Figure~\ref{fig:work_compare} reports the average circuit fidelity (10 runs with 1{,}000 shots each) of \design\ and three SDK-oriented baselines on three structured kernels under a uniform 0.2~MHz frequency detuning. \design\ (with Qwen30B) achieves near-unity fidelity across all kernels (above 0.99), whereas the baselines range from 0.83 to 0.97. Among the baselines, QCA exhibits the lowest fidelity, reflecting occasional inaccuracies in interpreting user intent and generating correct circuits, consistent with its role as an assistive tool for Qiskit APIs rather than an end-to-end synthesis system. Pennylang and QMAS achieve higher fidelity by reliably generating correct circuits for these relatively simple kernels. The evaluated kernels are intentionally simple to reduce circuit-generation difficulty across systems, so the remaining fidelity differences mainly reflect whether calibration is applied. Specifically, SDK-oriented methods generate circuits that are transpiled and executed without hardware-specific frequency correction, leaving them vulnerable to device-level detuning. In contrast, \design\ incorporates a calibration stage that parses backend constraint files and produces pulse-level OpenQASM code with drift compensation. This capability effectively mitigates detuning-induced errors. These results highlight a key advantage of \design: by integrating hardware-aware calibration into the code generation pipeline, it preserves execution fidelity under realistic hardware conditions, extending beyond circuit-level correctness to reliable physical execution.

\section{Conclusion}
In this paper, we present~\design, the first autonomous multi-agent framework for generating hardware-aligned OpenQASM programs. By integrating schema-aware planning, kernel-guided synthesis, and constraint-file–driven calibration, \design~enables reliable, end-to-end quantum program generation that adapts to device variability. Our evaluation across twelve representative kernels demonstrates substantial improvements over existing LLM-based methods, particularly on multi-kernel compositions. As NISQ systems continue to advance, we expect \design\ to serve as a foundation for scalable, hardware-aware quantum software automation. In future work, we plan to extend the framework to richer pulse-level control, dynamic circuit features, and broader quantum hardware platforms.

% We presented QAgent, an LLM-based multi-agent framework that automates OpenQASM programming from planning to calibration. By integrating schema-aware planning, hybrid coding strategies, and hardware-level calibration, QAgent achieves robust and accurate quantum code generation. Experiments show substantial improvements over static methods, \textcolor{red}{with above 98\% Pass@1 on all tasks.} These results highlight the potential of multi-agent LLM systems to bridge quantum algorithm design and hardware execution, paving the way toward practical and accessible quantum programming.

%\section*{Artifact Availability}
% The code associated with this work will be released upon acceptance.
%All code for this work will be released once the paper is accepted.

\section*{Acknowledgment}
This work was supported in part by 
NSF OAC-2417589 and NSF CNS-2143120. 
We thank the IBM Quantum Researcher \& Educators Program for their support of Quantum Credits.
Any opinions, findings and conclusions or recommendations expressed in this material are those of the authors and do not necessarily reflect the views of grant agencies or their contractors.

% \clearpage
%\bibliographystyle{ACM-Reference-Format}
\bibliographystyle{ieeetr}
\bibliography{agent}
\end{document}